\documentclass[a4paper,10pt,oneside,onecolumn,preprint,sort&compress]{elsarticle}

\usepackage{cite}
\usepackage{booktabs}
\usepackage{graphicx}
\usepackage{bm}
\usepackage{amsthm,amsmath,amssymb}
\usepackage{color}
\usepackage{enumitem}
\usepackage{caption}
\usepackage{subcaption}
\usepackage{mathtools}
\usepackage{dcolumn}
\usepackage{hyperref}
\usepackage[ruled]{algorithm2e}

\journal{Information Fusion}

\begin{document}

\begin{frontmatter}



\title{Distribution alignment based transfer fusion frameworks on quantum devices for seeking quantum advantages}


\author[1,2]{Xi He\corref{cor1}}
\ead{xihe@cuhk.edu.hk}

\author[1,3]{Feiyu Du}

\author[4]{Xiaohan Yu}

\author[5]{Yang Zhao}

\author[3]{Tao Lei}

\affiliation[1]{organization={Institute of Fundamental and Frontier Sciences, University of Electronic Science and Technology of China},
            city={Chengdu},
            postcode={610054}, 
            country={China}}
            
\affiliation[2]{organization={Department of Physics, The Chinese University of Hong Kong, Shatin, New Territories, Hong Kong, China},
            city={Hong Kong},
            country={China}}

\affiliation[3]{organization={Shaanxi Joint Laboratory of Artificial Intelligence, Shaanxi University of Science and Technology},
            city={Xi'an},
            postcode={710021}, 
            country={China}}

\affiliation[4]{organization={School of Computing, Macquarie University},
            city={Sydney},
            postcode={2109}, 
            country={Australia}}
            
\affiliation[5]{organization={College of Water Conservancy and Hydropower Engineering, Sichuan Agricultural University},
            city={Ya'an},
            postcode={625014}, 
            country={China}}

\cortext[cor1]{Corresponding author}

\begin{abstract}
The scarcity of labelled data is specifically an urgent challenge in the field of quantum machine learning (QML). Two transfer fusion frameworks are proposed in this paper to predict the labels of a target domain data by aligning its distribution to a different but related labelled source domain on quantum devices. The frameworks fuses the quantum data from two different, but related domains through a quantum information infusion channel. The predicting tasks in the target domain can be achieved with quantum advantages by post-processing quantum measurement results. One framework, the quantum basic linear algebra subroutines (QBLAS) based implementation, can theoretically achieve the procedure of transfer fusion with quadratic speedup on a universal quantum computer. In addition, the other framework, a hardware-scalable architecture, is implemented on the noisy intermediate-scale quantum (NISQ) devices through a variational hybrid quantum-classical procedure. Numerical experiments on the synthetic and handwritten digits datasets demonstrate that the variational transfer fusion (TF) framework can reach state-of-the-art (SOTA) quantum DA method performance. 
\end{abstract}

%


\end{frontmatter}


\section{Introduction}
\label{sec:introduction}
Quantum computation is demonstrated to take advantage of the properties of quantum mechanics to enhance the performance of computational process~\citep{nilsen&chuang,shor1994algorithms,grover1996fast,harrow2009quantum}. Based on the quantum entanglement and superposition operations, the quantum computing is merged with machine learning (ML) resulting in the field of quantum machine learning (QML)~\citep{QML}. On the one hand, the application scope of ML can be extended by the QML with performance advantages~\citep{huang2021power,huang2022provably,huang2020predicting}; on the other hand, the QML methods have effectively promoted the performance in dealing with a variety of tasks compared to the classical ML methods~\citep{QML_book,quantum_kernel1,quantum_kernel2}. At present, the QML has developed into a comprehensive and systematic research field. In terms of the types of learning task, both the discriminative~\citep{QSVM,QNN,QKNN,QDF,QLR,QLDA} and generative tasks~\citep{QPCA,QLLE,QAE1,QAE2,QVAE,QRBM1,QRBM2,QGAN1,QGAN2,QGAN3,QGAN4,QGAN5,QGAN6} can be efficiently accomplished with quantum advantages. In addition, the QML methods for different quantum devices such as the universal quantum computer and the NISQ devices are proposed for seeking quantum advantages under different circumstances. The goal of the QML is not to defeat the classical ML, but to enlarge the ML community for further exploration in computing science. 

As the increase of the scale of available data, the ML method has to deal with learning tasks from different domains simultaneously with domain shift. In addition, data from specific domains such as quantum data are expensive and unlabelled. Traditional ML methods, which obey the independent and identically distributed (IID) assumption, may not be able to handle this situation. In the field of ML, the transfer learning (TL) devotes to accomplish the task in an unprocessed target domain with the acquired knowledge of a processed source domain~\citep{pratt1993discriminability,pan2009survey}. As an important sub-realm of TL, domain adaptation (DA) attempts to predict the labels of the unprocessed target domain data set based on the labelled source domain data set. 

As a crucial method of DA, the distribution domain adaptation (DDA) assumes that the distributions of the labelled source domain data and the unlabelled target domain data are generated from different data distributions. The goal of the DDA is to predict the target labels of an unprocessed dataset by minimizing the distribution discrepancies between the labelled source domain data and the unlabelled target domain data together~\citep{TCA,JDA,BDA}. In order to achieve a competitive adaptation performance, the distribution adaptation considers to approximate both the marginal and the conditional distributions of the two domains. However, the classical distribution adaptation algorithm can be time-consuming with the increase of the number and dimension of the given data. In addition, the computational complexity of the classical DDA can be prohibited in the quantum data scenario.

In addition to the encoding from classical data, quantum data are mostly generated from the evolution of quantum systems. On the one hand, the quantum data and the corresponding labels are rare and expensive to achieve; on the other hand, quantum tomography operations which are extremely time-consuming are necessary for classical ML algorithm to handle the domain shift between different quantum data domains.

To deal with this dilemma, different types of quantum transfer learning methods are proposed. Ref.~\citep{mari2019transfer} proposed a framework for achieving the procedure of TL through the variational quantum neural networks. They discussed different extensive TL scenarios and presented many representative examples to demonstrate the efficiency of the quantum transfer learning (QTL). Furthermore, for quantum domain adaptation (QDA), a quantum marginal distribution adaptation algorithm, namely the quantum transfer component analysis (QTCA)~\citep{QTCA}, is proposed to achieve the domain adaptation procedure by approximating the marginal distributions of the source and target domain. In addition, the quantum feature alignment method~\citep{QSA,QCORAL,QCDA,QKSA}, a representative type of quantum subspace learning algorithm, can be implemented to achieve the procedure of DA on both the universal and NISQ devices.   

The works above only consider to align the marginal probability distributions of the source and target domain, or roughly the data matrix. The intra-class feature divergence between the two domains are neglected, which may harm the transfer performance. In this paper, to achieve the procedure of TL efficiently, the quantum information infusion channel $\mathcal{E}$ is designed to fuses the quantum source state $\rho_{s}$ and the quantum target state $\rho_{t}$. Combined with quantum measurement operations and classical post-processing, the target labels can be predicted by aligning both the marginal and conditional probability distributions as depicted in Fig.~\ref{Fig:1}. Concretely, two distribution alignment based TF frameworks are implemented on the universal and NISQ quantum devices respectively. The QBLAS-based TF framework can accomplish the procedure of DA on a universal quantum computer with quadratic quantum speedup compared to the classical TL methods. The variational quantum TF (VQTF) framework can align the distributions between the source and target domain through a variational hybrid quantum-classical procedure on NISQ devices. The quantum advantages of the quantum TF (QTF) frameworks in our work mainly comes from the quantum superposition encoding and the quantum kernel method. In addition to the theoretical computational complexity analysis, numerical experiments on the synthetic dataset for the quantum data scenario and the handwritten digits dataset for the classical data scenario are implemented, demonstrating that the VQTF can achieve SOTA quantum DA performance.

The remainder of this paper is arranged as follows. In section~\ref{sec:pre}, the procedure of domain distribution adaptation (DDA) and basic quantum computing operations are briefly overviewed. In the next, the implementation of the QBLAS-based TF is proposed in section~\ref{subsec:QBLAS-based_TF}. Subsequently, the implementation of the VQTF on the NISQ devices is provided in section~\ref{subsec:VQTF}. To demonstrate the feasibility and effectiveness of the NISQ implementation, two numerical experiments are presented in section~\ref{sec:numerical experiments}. Finally, the whole contents are concluded in section~\ref{sec:conclution}.

\section{Preliminary}
\label{sec:pre}
\subsection{Domain distribution adaptation (DDA)}
\label{subsec:DDA}
\begin{figure*}
	\centering
	\includegraphics[width=0.6\textwidth]{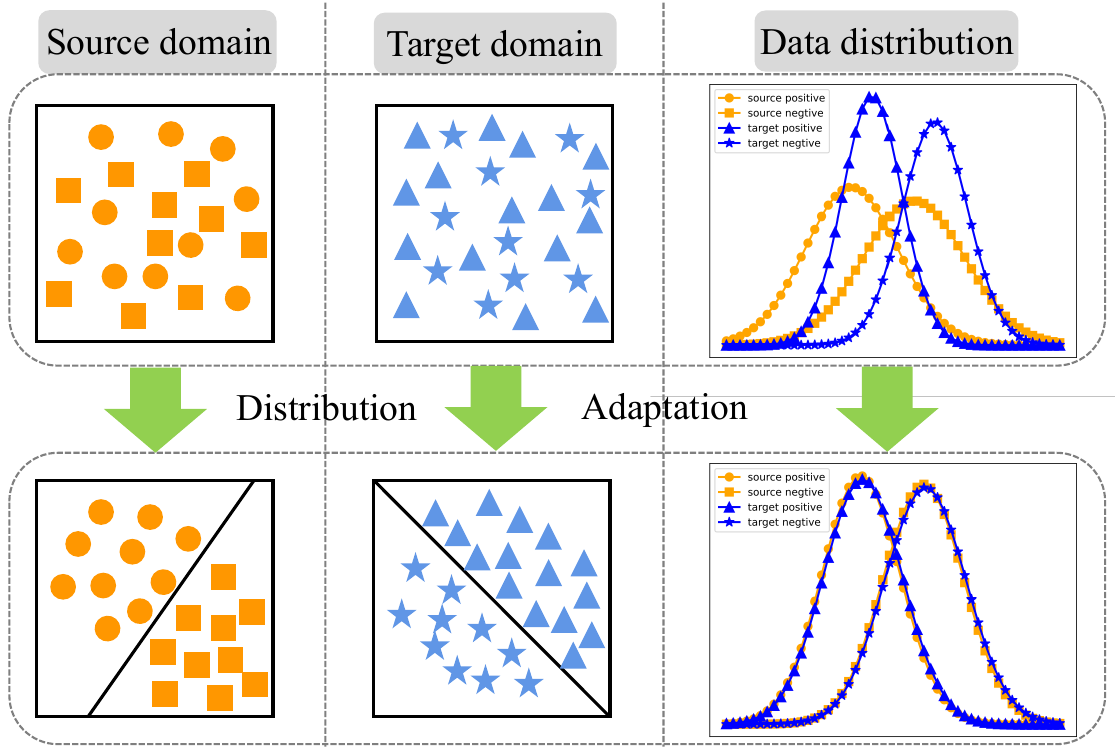}
	\caption{The schematic diagram of the procedure of the DDA. The source domain data $X_{s}$ and the target domain data $X_{t}$ are different in both the marginal and conditional distributions. The DDA attempts to align the distributions of the two domains to achieve the procedure of TL.}
	\label{Fig:1}
\end{figure*}

Distribution domain adaptation (DDA), a crucial method of DA, is designed to align both the marginal and conditional distributions of the labelled source domain with those in the unlabelled target domain, so that a classifier can be trained and applied to predict the target labels. Given a labeled source domain $\mathcal{D}_{s} = \{X_{s}, P(X_{s})\}$ and an unlabelled target domain $\mathcal{D}_{t} = \{ X_{t}, P(X_{t}) \}$ where the source domain data $X_s = (x_{1}^{(s)}, \cdots, x_{i}^{(s)}, \cdots, x_{n_{s}}^{(s)}) \in \mathbb{R}^{D \times n_{s}}$ with labels $Y_{s} = \{ y_{i}^{(s)} \}_{i=1}^{n_{s}} \in \mathcal{Y}_{s}$; the target domain data $X_{t} = (x_{1}^{(t)}, \cdots, x_{j}^{(t)}, \cdots, x_{n_{t}}^{(t)}) \in \mathbb{R}^{D \times n_{t}}$. Assume that both the marginal and the conditional distributions of the two domains are different, namely $P(X_s) \neq P(X_t)$ and $P(Y_s|X_s) \neq P(Y_t|X_t)$. But the two domains share the same feature and label space, namely $\mathcal{X}_s = \mathcal{X}_t$ and $\mathcal{Y}_s = \mathcal{Y}_t$. The goal of DDA is to minimize the distribution discrepancies between $P(X_{s})$ and $P(X_{t})$; $P(Y_{s} | X_{s})$ and $P(Y_{t} | X_{t})$ respectively as schematically depicted in Fig.~\ref{Fig:1}. Specifically, this method maps the original data to a higher-dimensional space by the kernel trick, and subsequently projects the data to a lower-dimensional space by a transformation matrix $W$. After the procedure of DDA, a classifier can be trained on the aligned labelled source domain and subsequently applied to the target domain to predict the target labels. 

Overall, the procedure of DDA mainly contains the marginal distribution adaptation and the conditional distribution adaptation. Let the input data matrix $X = (x_{1}^{(s)}, \cdots, x_{n_{s}}^{(s)}, x_{1}^{(t)}, \cdots, x_{n_{t}}^{(t)}) = (x_{1}, \cdots, x_{i}, \cdots, x_{n}) \in \mathbb{R}^{D \times n}$ where $n = n_{s} + n_{t}$. The maximum mean discrepancy (MMD) is adopted to measure the discrepancies between the distributions of the two domains as follows
\begin{eqnarray}
	dist(P_{s}, P_{t}) &= \left \vert \frac{1}{n_{s}}\sum_{i=1}^{n_{s}} \phi(x_i^{(s)}) - \frac{1}{n_{t}}\sum_{j=1}^{n_{t}} \phi(x_j^{(t)}) \right \vert^{2} = \mathrm{tr} \left( KL_{0}\right),
	\label{eq:MMD}
\end{eqnarray}
where $\phi$ is the kernel map; $\mathrm{tr}(\cdot)$ represents the trace operation; the kernel matrix $K = [
	\phi(X_{s}) \ \phi(X_{t}))
]^{T} [
	\phi(X_{s}) \ \phi(X_{t}))
] = (k_{1}, \cdots, k_{n}) \in \mathbb{R}^{n \times n}$; $L_{0} = l_{0}^{T} l_{0}$ with $l_{0} = (1 / n_{s} \textbf{1}_{n_{s}}, -1 / n_{t} \textbf{1}_{n_{t}})$; $\textbf{1}_{n} = (1, \cdots, 1)_{n}$ represents a $n$-dimensional vector whose elements are all equal to one. Decompose the kernel matrix $K = (KK^{-1/2})(K^{-1/2}K)$, the marginal distribution adaptation can be achieved by minimizing
\begin{eqnarray}
	dist(P(X_{s}), P(X_{t})) &=& \mathrm{tr}((KK^{-1/2} \widetilde{W})(\widetilde{W}^{T}K^{-1/2}K)L_{0})  \nonumber \\
	&=& \mathrm{tr}(KWW^{T}KL_{0}) \nonumber \\ 
	&=& \mathrm{tr}(W^{T}KL_{0}KW)
	\label{eq:MDA distance}
\end{eqnarray} 
where the transformation matrix $W = K^{-1/2}\widetilde{W}$~\citep{QTCA}.

Similarly, the conditional distribution adaptation of the two domains can be achieved as follows. The conditional distribution distance can be defined as
\begin{equation}
    dist(P(Y_s | X_s), P(Y_t | X_t)) = \mathrm{tr} \left( W^{T} K L_{c} K W \right)
    \label{eq:CDA distance}
\end{equation}
where $L_{c} = l_{c}^{T}l_{c}$ with $l_{c} = (1 / n_{s}^{(c)} \textbf{1}_{n_{s}^{(c)}}, -1/n_{t}^{(c)} \textbf{1}_{n_{t}^{(c)}})$; $n_s^{(c)}$, $n_{t}^{(c)}$ are the number of samples which come from the source and target domain of class $c$ respectively with $c \in \{1, \cdots, C \}$ after the pseudo target label prediction.

Integrate the two distances above resulting in the total distance to estimate the divergence between the two domains as
\begin{eqnarray}
    dist(\mathcal{D}_s, \mathcal{D}_t) & \approx & (1-\kappa)dist(P(X_{s}), P(X_{t})) \nonumber \\
    & \ & + \kappa dist(P(Y_{s}|X_{s}), P(Y_{t}|X_{t}))
    \label{eq:distribution adaptation distance}
\end{eqnarray}
with a balance factor $\kappa$.

Therefore, the objective function of the distribution function can be reduced to 
\begin{eqnarray}
	&& \min \mathrm{tr}(W^{T}K((1-\kappa)L_0 + \kappa\sum_{c=1}^{C}L_{c})KW) + \mu \Vert W \Vert_{F}^{2} \\ 
    && s.t.\ W^{T}KMKW = I,\ 0 \leq \mu \leq 1 
    \label{eq:objective function}
\end{eqnarray}
where the matrix $M = I_{n} - 1 / n \textbf{1}_{n}^{T} \textbf{1}_{n}$; $\Vert \cdot \Vert_{F}$ is the Frobenius norm. Subsequently, define the Lagrange function and take the derivative of $W$ equal zero resulting in
\begin{equation}
    (KL_{Q}K + \mu I)W = KMKW\Omega
    \label{eq:eigen equation}
\end{equation}
where $L_{Q} = (1-\kappa)L_0 + \kappa\sum_{c=1}^{C}L_c$; $\Omega$ is a diagonal matrix whose diagonal elements are respectively corresponding Lagrange multiplier parameters.

In the end, the weight matrix $W$ can be constructed by extracting $d$ eigenvectors which are in correspondence to the $d$ smallest eigenvalues of the matrix $G = (KL_{Q}K + \mu I)^{-1}KMK$. A classifier $f$ is trained on $\{ \hat{x}^{(s)}_{i} = W^{T} x_{i}, y_{i}^{(s)} \}_{i=1}^{n_{s}}$ and utilized to predict the pseudo target labels $\{ \tilde{y}_{j}^{(t)} = f(W^{T} x_{n_{s} + j}) \}_{j=1}^{n_{t}}$. Iterate the pseudo target label prediction and objection function minimization until convergence resulting in the optimal transformation matrix $W_{\ast}$ and the final target labels $Y_{t}$. 

\subsection{Quantum computation}
\label{subsec:QC}
Based on the principles of quantum mechanics, quantum computation and quantum information method attempts to deal with quantum data efficiently. The basic quantum concepts and operations are overviewed in this section.

\subsubsection{Quantum qubits}
\label{subsubsec:qubit}
Compared with classical computation, quantum computation uses qubit as the basic data unit. A single qubit can be represented by a two-dimensional vector in the Hilbert space. Based on the Dirac notation and a set of orthogonal basis, a single-qubit state 
\begin{equation}
    | \psi \rangle = \alpha | 0 \rangle + \beta | 1 \rangle,
    \label{eq:single_qubit}
\end{equation}
where $\alpha$, $\beta$ are the probability amplitude with $\vert \alpha \vert^{2} + \vert \beta \vert^{2} = 1$. Specifically, after the quantum measurement operation on this qubit, $| \psi \rangle$ collapses to $| 0 \rangle$ with probability $\vert \alpha \vert^{2}$ or to $| 1 \rangle$ with probability $\vert \beta \vert^{2}$.

For a $n$-qubit quantum state
\begin{equation}
    | \psi_{n} \rangle = \alpha_{0 0 \cdots 0 0} | 0 \cdots 0 \rangle + \alpha_{0 0 \cdots 0 1} | 0 0 \cdots 0 1 \rangle + \cdots + \alpha_{1 1 \cdots 1 1} | 1 1 \cdots 1 1 \rangle,
    \label{eq:multi-qubit}
\end{equation}
where $\vert \alpha_{0 0 \cdots 0 0} \vert^{2} + \vert \alpha_{0 0 \cdots 0 1} \vert^{2} + \cdots + \vert \alpha_{1 1 \cdots 1 1} \vert^{2} = 1$ for $n \geq 2$.

In addition to the representation above, a quantum state can also be represented by a density operator $\rho = | \psi \rangle \langle \psi |$.

\subsubsection{Quantum operations}
\label{subsubsec:operation}
Quantum circuits utilizes a series of quantum gates to implement the procedure of quantum computation. The most basic quantum gates are the single-qubit gates and the controlled unitary gate. The process of performing quantum operations $U$ on the quantum gate $| \psi_{init} \rangle$ can be schematically as
\begin{equation}
    | \psi_{final} \rangle = U | \psi_{init} \rangle.
    \label{eq:quantum_operation}
\end{equation}

The common quantum gates is depicted as Fig.**. Specifically, the Pauli-X, Y, Z and the Hadamard gates are
\begin{equation}
    X = \begin{bmatrix} 0 & 1 \\ 1 & 0 \end{bmatrix}, \
    Y = \begin{bmatrix} 0 & -i \\ i & 0 \end{bmatrix}, \
    Z = \begin{bmatrix} 1 & 0 \\ 0 & -1 \end{bmatrix}, \
    H = \frac{1}{\sqrt{2}}\begin{bmatrix} 1 & 1 \\ 1 & -1 \end{bmatrix},
    \label{eq:XYZH}
\end{equation}
respectively.
The controlled unitary gate
\begin{equation}
    C-U = | c \rangle | t \rangle \rightarrow | c \rangle U^{c} | t \rangle \Leftrightarrow \begin{bmatrix}
        I & 0 \\ 0 U
    \end{bmatrix}
\end{equation}
where $c$ is the controlled qubit; $U$ is the unitary operation to perform on the qubit $| t \rangle$, which depends on the state of $| c \rangle$. 

Combined with the basic quantum gates above, the quantum circuits can achieve arbitrary complex computation such as the quantum Fourier transform~\citep{nilsen&chuang}.

\begin{figure*}
	\centering
	\begin{subfigure}{0.2\textwidth}
	\centering
	\includegraphics[width=\textwidth]{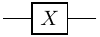}
	\caption{}
	\label{Fig.2a}
	\end{subfigure}	
	\begin{subfigure}{0.2\textwidth}
	\centering
	\includegraphics[width=\textwidth]{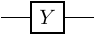} 
	\caption{}
	\label{Fig.2b}
	\end{subfigure}
    \begin{subfigure}{0.2\textwidth}
	\centering
	\includegraphics[width=\textwidth]{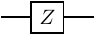} 
	\caption{}
	\label{Fig.2c}
	\end{subfigure}
	\\
	\begin{subfigure}{0.2\textwidth}
	\centering
	\includegraphics[width=\textwidth]{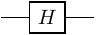}
	\caption{}
	\label{Fig.2d}
	\end{subfigure}	
	\begin{subfigure}{0.2\textwidth}
	\centering
	\includegraphics[width=\textwidth]{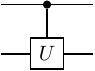} 
	\caption{}
	\label{Fig.2e}
	\end{subfigure}
    \begin{subfigure}{0.15\textwidth}
	\centering
	\includegraphics[width=\textwidth]{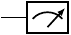} 
	\caption{}
	\label{Fig.2f}
	\end{subfigure}
	\caption{The schematic notaion of the basic quantum gates. (a)Pauli-X gate; (b)Pauli-Y gate; (c)Pauli-Z gate; (d)Hadamard gate; (e)Controlled-$U$ gate; (f)Quantum measurement.}
	\label{Fig:6}
\end{figure*} 

\section{Distribution alignment based quantum transfer fusion frameworks}
\label{sec:DAQTF}
\begin{figure*}[tb]
	\centering
	\includegraphics[width=0.9\textwidth]{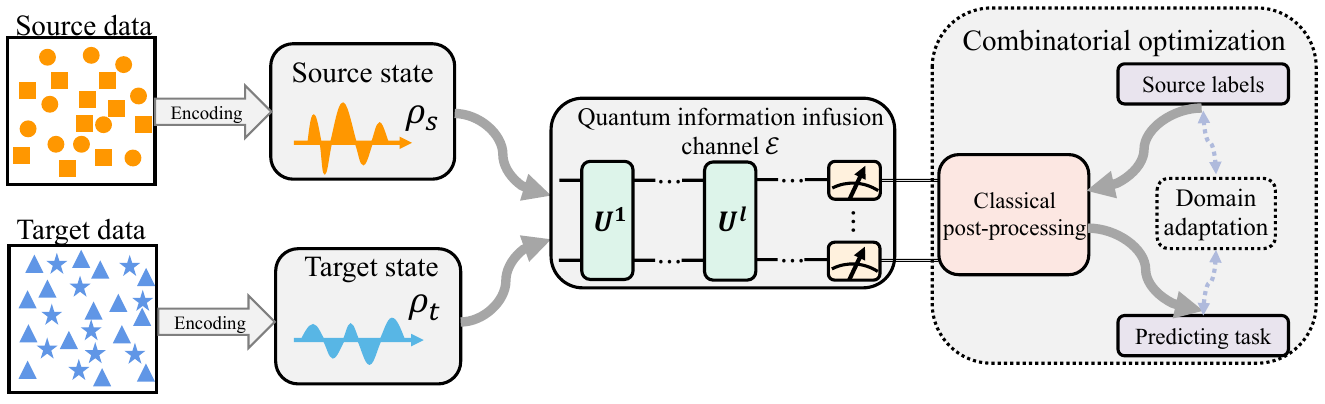}
	\caption{The schematic diagram of the distribution alignment based quantum transfer fusion framework.}
	\label{Fig:3}
\end{figure*}

As depicted in Fig.~\ref{Fig:3}, given data from the source and target domain with different marginal and conditional distributions, the corresponding quantum state $\rho_{s}$ and $\rho_{t}$ can be prepared by data encoding. Subsequently, the quantum information infusion channel $\mathcal{E}$ fuses the data features from the two domains and produces the results by the quantum measurement operations on all qubit registers. Through classical post-processing, the target labels prediction can be achieved with quantum advantages. 

\subsection{Quantum basic linear algebra subroutines based transfer fusion framework (QBLAS-based TF)}
\label{subsec:QBLAS-based_TF}
\begin{figure*}[t]
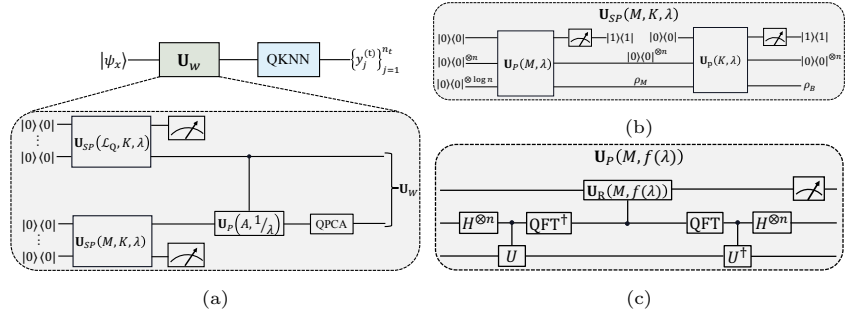

	\centering
	\begin{subfigure}{0.45\textwidth}
	\centering
	\includegraphics[width=\textwidth]{Fig4a.pdf}
	\caption{}
	\label{Fig.4a}
	\end{subfigure}	
	\begin{subfigure}{0.45\textwidth}
	\centering
	\includegraphics[width=\textwidth]{Fig4b.pdf}
	\caption{}
    \label{Fig.4b}
	\includegraphics[width=\textwidth]{Fig4c.pdf}
	\caption{}
	\label{Fig.4c}
	\end{subfigure}
	\caption{The corresponding quantum circuits of the QBLAS-based TF. (a) The quantum circuits for predicting the target labels through the procedure of QBLAS-based TF where $U_{w}$ is the quantum information fusion channel with $QPCA$ represents the quantum principal component analysis operation; (b) The quantum circuits of $\textbf{U}_{SP}(M, K, \lambda)$; (c) The quantum circuits of $\textbf{U}_{P}(M, f(\lambda))$ where $QFT$ and $QFT^{\dagger}$ are the quantum Fourier transform and its inverse respectively.}
	\label{Fig:4}
\end{figure*}

In this section, the procedure of the QTF is implemented on the universal quantum computer as depicted in Fig.~\ref{Fig:4}. After the clarification of the essential assumption, the QBLAS-based quantum information channel $U_{w}$ can be prepared in section~\ref{subsubsec:QBLAS_channel}. Subsequently, the pseudo label predictor is presented in section~\ref{subsubsec:QLP}. Finally, the DDA can be implemented on the universal quantum device by the quantum basic linear algebra subroutines (QBLAS) with quantum quadratic speedup theoretically guaranteed in section~\ref{subsubsec:QBLAS-based TF}. 

\subsubsection{Implementation of the QBLAS-based information fusion channel}
\label{subsubsec:QBLAS_channel}
For the original data given in form of classical vectors, we assume that the elements of $X_{s}$ and $X_{t}$ are both accessible in time $\mathcal{O}(\rm{poly}(\log n))$ with the quantum random access memory (qRAM)~\citep{QRAM}. If quantum data are given originally, they can be directly processed without amplitude encoding. The given data $X$ can be represented by the quantum state
\begin{equation}
	| \psi_{X} \rangle = \sum_{i=1}^{n} \sum_{m=1}^{D} x_{mi} | i \rangle | m \rangle,
	\label{eq:data state}
\end{equation}
where $x_{mi}$ is the normalized element of $X$. The kernel matrix $K$ can be constructed by the kernel method. The kernel feature map $\phi: x \rightarrow | \phi(x) \rangle$ can be specified manually with different selections~\citep{quantum_kernel1,quantum_kernel2}. Based on the quantum kernel tricks, the state
\begin{equation}
	| \phi(X) \rangle = \sum_{i=1}^{n} | i \rangle | \phi(x_{i}) \rangle
	\label{eq:kernel state}
\end{equation}
can be obtained. Hence, the kernel matrix $K$ can be represented by the kernel state $\rho_{K} = \mathrm{tr}_{i} \{ | \phi(X) \rangle \langle \phi(X) | \}$ in $\mathcal{O}( \kappa_{K} \log n / \epsilon^{3})$ where $\kappa_{K}$ is the condition number of $K$; $\epsilon$ is the error parameter. In addition, after encoding the matrix $M$ to the state $\rho_{M}$, $\rho_M$ can be decomposed in the eigenbasis of $K$ resulting in 
\begin{equation}
	\rho_{0} = \sum_{i, j=1}^{n} \langle u_{i} | \rho_{M} | u_{j} \rangle | u_{i} \rangle \langle u_{j} |,
	\label{eq:M}
\end{equation}
where $\{ u_{i} \}_{i=1}^{n}$ represent the eigenvectors of $K$.

Perform the phase estimation (PE)~\citep{nilsen&chuang} on $\rho_{0}$, the quantum state
\begin{equation}
	\rho_{1} = \sum_{i,j=1}^{n} \langle u_{i} | \rho_{M} | u_{j} \rangle | \lambda_{i} \rangle \langle \lambda_{j} | \otimes | u_{i} \rangle \langle u_{j} |,
	\label{eq:rho_1}
\end{equation}
where $\{ \lambda_{i} \}_{i=1}^{n}$ are the eigenvalues of $K$.

Then, the controlled rotation operation $C-R_{y}(2\arcsin (\gamma_1 \lambda_{i}))$ is applied on a newly added ancilla resulting in 
\begin{equation}
	\rho_2 = \sum_{i,j=1}^{n} \langle u_{i} | \rho_{M} | u_{j} \rangle | \lambda_{i} \rangle \langle \lambda_{j} | \otimes | u_{i} \rangle \langle u_{j} | \otimes | \psi_{1i} \rangle \langle \psi_{1j} |,
	\label{eq:rho_2}
\end{equation}
with $| \psi_{1i} \rangle = \sqrt{1 - \gamma_{1}^2 \lambda_{i}^2} | 0 \rangle + \gamma_{1} \lambda_{i} | 1 \rangle$ and $\gamma_{1}$ is a constant.

Uncompute the eigenvalue register and measure the ancilla to be $| 1 \rangle \langle 1 |$ resulting in the state
\begin{equation}
	\rho_{B} = \frac{1}{\sqrt{P(B)}} \sum_{i, j=1}^{n} \gamma_{1}^{2} \lambda_{i} \lambda_{j}^{\ast} \langle u_{i} | \rho_{M} | u_{j} \rangle | u_{i} \rangle \langle u_{j} |
	\label{eq:rho_{B}}
\end{equation}
proportional to the matrix $B = KMK$ in $\mathcal{O}(\kappa_{K}^{4} \log n / \epsilon^{3})$ to error $\epsilon$ where $P(B) = \sum_{i, j=1}^{n} \vert \gamma_{1}^{2} \lambda_{i} \lambda_{j}^{\ast} \vert^{2}$.

Similarly, the state $\rho_{L_{0}}$ representing the matrix $L_{0}$ can be decomposed into
\begin{equation}
	\rho_{3} = \sum_{i, j=1}^{n} \langle u_{i} | \rho_{L_{0}} | u_{j} \rangle | u_{i} \rangle \langle u_{j} |.
	\label{eq:L_0}
\end{equation}

Subsequently, the PE and the $C-R_{y}(2\arcsin(\gamma_{2} \lambda_{i}))$ rotation operation can be performed on $\rho_{3}$ sequentially with a newly added register yielding
\begin{equation}
	\rho_{4} = \sum_{i,j=1}^{n} \langle u_{i} | \rho_{L_{0}} | u_{j} \rangle | \lambda_{i} \rangle \langle \lambda_{j} | \otimes | u_{i} \rangle \langle u_{j} | \otimes | \psi_{2i} \rangle \langle \psi_{2j} |
	\label{eq:rho_4}
\end{equation}
with $| \psi_{2i} \rangle = \sqrt{1 - \gamma_{2}^2 \lambda_{i}^2} | 0 \rangle + \gamma_{2} \lambda_{i} | 1 \rangle$ and $\gamma_{2}$ is a constant.

After uncomputing the eigenvalue register and measuring the ancilla register to be $| 1 \rangle \langle 1 |$, the quantum state
\begin{equation}
	\rho_{A} = \frac{1}{\sqrt{P(A)}} \sum_{i, j=1}^{n} \gamma_{2}^{2} \lambda_{i} \lambda_{j}^{\ast} \langle u_{i} | \rho_{L_{0}} | u_{j} \rangle | u_{i} \rangle \langle u_{j} |
	\label{eq:A}
\end{equation}
representing the matrix $A = K L_{0} K$ can be obtained in $\mathcal{O}(\kappa_{K}^{4} \log n / \epsilon^{3})$ to error $\epsilon$ with $P(A) =\sum_{i, j=1}^{n} \vert \gamma_{2}^{2} \lambda_{i} \lambda_{j}^{\ast} \vert^{2}$.

\subsubsection{Quantum label predictor}
\label{subsubsec:QLP}
After the state preparation, a label predictor trained on the aligned source domain data for predicting the target labels is specifically invoked for pseudo label prediction. In the QBLAS-based TF framework, the quantum $k$-nearest neighbors algorithm (QKNN)~\citep{QKNN} is adopted after the distribution alignment. The label of the specified target data point can be determined as the label of the nearest data point by computing and searching all the distances among the data points for the minimum. The procedure of applying the QKNN for pseudo target label prediction is presented as follows.

(1) Encode the aligned target data point $\hat{x}_{j}^{(t)}$ as the quantum state
\begin{equation}
	| \alpha \rangle = \frac{1}{\sqrt{D}} \sum_{m=1}^{D} | m \rangle \left ( \sqrt{1 - \hat{x}_{mj}^{(t)2}} | 0 \rangle + \hat{x}_{mj}^{(t)} | 1 \rangle \right ) | 0 \rangle,
	\label{eq:alpha state}
\end{equation}
and the aligned source domain data as the quantum state
\begin{equation}
	| \beta \rangle = \frac{1}{\sqrt{n_{s}}} \sum_{i=1}^{n_s} | i \rangle \frac{1}{\sqrt{D}} | m \rangle | 0 \rangle \left ( \sqrt{1 - \hat{x}_{mi}^{(s)2}} | 0 \rangle + \hat{x}_{mi}^{(s)} | 1 \rangle \right ).
	\label{eq:beta state}
\end{equation}

(2) Perform the swap test on $| \alpha \rangle$ and $| \beta \rangle$ resulting in the quantum state 
\begin{equation}
	| \gamma \rangle = \frac{1}{\sqrt{n_{s}}} \sum_{i=1}^{n_{s}} | i \rangle \left ( \sqrt{1 - d(\hat{x}_{j}^{(t)}, \hat{x}_{i}^{(s)})} | 0 \rangle + \sqrt{d(\hat{x}_{j}^{(t)}, \hat{x}_{i}^{(s)})} | 1 \rangle \right ), 
	\label{eq:gamma state}
\end{equation}
where $d(\hat{x}_{j}^{(t)}, \hat{x}_{i}^{(s)}) = \sqrt{2 - 2 \vert \langle \hat{x}_{j}^{(t)} | \hat{x}_{i}^{(s)} \rangle \vert}$ is the Euclidean distance between the two data points.

(3) Through the amplitude estimation algorithm, the state
\begin{equation}
	| \sigma \rangle = \frac{1}{n_{s}} \sum_{i=1}^{n_{s}} | i \rangle | d(\hat{x}_{j}^{(t)}, \hat{x}_{i}^{(s)}) \rangle
	\label{eq:sigma state}
\end{equation}
can be obtained. Ultimately, the pseudo label of $x_{j}^{(t)}$ is exactly the same as the label of the source domain data point with the minimum distance by the D$\ddot{u}$rr's algorithm in time $\mathcal{O}(\sqrt{kn_{s}})$.

\subsubsection{QBLAS-based TF}
\label{subsubsec:QBLAS-based TF}
After the preparation of the quantum states and the pseudo target label predictor, the procedure of the QBLAS-based TF is presented as follows.

(1) Prepare the quantum state $\rho_{A}$ proportional to the matrix $A = K((1-\kappa)L_0 + \kappa\sum_{c=1}^{C}L_c)K + \mu I$ as presented in section~\ref{subsubsec:QBLAS_channel}.

(2) Decompose $\rho_{B}$ in the eigenbasis of $\rho_{A}$ as 
\begin{equation}
	\rho_{\hat{A}} = \sum_{i, j=1}^{n} \langle u_{A_{i}} | \rho_{B} | u_{A_{j}} \rangle | u_{A_{i}} \rangle \langle u_{A_{j}} |,
	\label{eq:rho_5}
\end{equation}
where $\{ u_{A_{i}} \}_{i=1}^{n}$ are the eigenvectors of $\rho_{A}$.

(3) Perform the QPE on $\rho_{\hat{A}}$ and the controlled $R_{y}(2\arcsin(\gamma_{3} \lambda_{A_{i}}^{-1/2}))$ on a newly added ancilla resulting in 
\begin{equation}
	\rho_{\tilde{A}} = \sum_{i, j=1}^{n} \langle u_{A_{i}} | \rho_{B} | u_{A_{j}} \rangle | \lambda_{A_{i}} \rangle \langle \lambda_{A_{j}} | \otimes | u_{A_{i}} \rangle \langle u_{A_{j}} | \otimes | \psi_{3i} \rangle \langle \psi_{3j} |,
	\label{eq:rho_6}
\end{equation}
where $| \psi_{3i} \rangle = \sqrt{1 - \gamma_{3}^{2} / \lambda_{A_{i}}} | 0 \rangle + \gamma_{3} \lambda_{A_{i}}^{-1/2} | 1 \rangle$.

(4) Uncompute the eigenvalue register and measure the ancilla register to be $| 1 \rangle \langle 1 |$ to obtain
\begin{equation}
	\rho_{G} = \frac{1}{\sqrt{P(G)}} \sum_{i, j=1}^{n} \frac{\gamma_{3}^{2}}{\sqrt{\lambda_{A_{i}}\lambda_{A_{j}}^{\ast}}}\langle u_{A_{i}} | \rho_{B} | u_{A_{j}} \rangle | u_{A_{i}} \rangle \langle u_{A_{j}} |,
	\label{eq:rho_G}
\end{equation}
which is proportional to the matrix $G$ in $\mathcal{O}(\kappa_{A}^{4} \log n / \epsilon^{3})$ to error $\epsilon$ with $P(G) = \vert \sum_{i, j=1}^{n} \gamma_{3}^{2} (\lambda_{A_{i}} \lambda_{A_{j}}^{\ast})^{-1/2} \vert^{2}$.

(5) Invoke the qPCA~\citep{QPCA} to extract the $d$ largest eigenvalues and corresponding eigenvectors of the matrix $\hat{G} = \eta I - \rho_{G}$ in time $\mathcal{O}(\sqrt{d})$ and subsequently obtain the weight matrix $W$ where $\eta$ is a constant.

(6) Train the classifier $f_{K}$ on $\{ \hat{x}^{(s)}_{i}, y_{i}^{(s)} \}_{i=1}^{n_{s}}$; update the pseudo target labels $\{ \tilde{y}_{j}^{(t)} \}_{j=1}^{n_{t}}$ and the matrix $L_{c}$.

(7) Iterate step (1) to step (6) until convergence resulting in the optimal classifier $f_{K\ast}$ and the target labels $\{ y_{j}^{(t)} \}_{j=1}^{n_{t}}$.

Ultimately, the procedure of QBLAS-based TF framework can be achieved by the above iterative steps. In addition, the pseudo code of this framework is provided in Algorithm~\ref{alg1}. The corresponding quantum circuits are presented in Fig.~\ref{Fig:4}.

Overall, the QBLAS-based TF method can be implemented on a universal quantum computer with the computational complexity in $O(\sqrt{k n_{s}})$. The quadratic quantum speedup comes both directly from the quantum superposition of the data encoding, and indirectly from the sparsity after mapping the data to the quantum kernel space.

\begin{algorithm}[t]
	\caption{QBLAS-based TF framework}
	\KwIn{Source domain data $X_{s} = \{ x_{i}^{(s)} \}_{i=1}^{n_{s}}$ with labels $Y_{s} = \{ y_{i}^{(s)} \}_{i=1}^{n_{s}}$, target domain data $X_{t} = \{ x_{j}^{(t)} \}_{j=1}^{n_{t}}$}
	\KwOut{Target labels $\{ y_{j}^{(t)} \}_{j=1}^{n_{t}}$}
	\emph{step 1}: Prepare the kernel matrix $K$, $\rho_{B}$, $\rho_{A}$ as presented in section~\ref{subsubsec:QBLAS_channel}. \\
	\emph{step 2}: \\
	\Repeat{Convergence}{
	(1) Solve the weight matrix $W$ as in Eq.~\eqref{eq:rho_5} to Eq.~\eqref{eq:rho_G}; \\
	(2) Invoke the qPCA to construct the weight matrix $W$; \\
	(3) Invoke the quantum classifier $f_{K}$ in section~\ref{subsubsec:QLP} to predict the pseudo target labels $\{ \tilde{y}_{j}^{(t)} \}_{j=1}^{n_{t}}$; \\
	(4) Update the matrix $L_{c}$.
	}
	Return the target labels $Y_{t} = \{ y_{j}^{(t)} \}_{j=1}^{n_{t}}$.
	\label{alg1}
\end{algorithm}

\subsection{Variational quantum transfer fusion framework (VQTF)}
\label{subsec:VQTF}
Although the procedure of information fusion can be implemented on a universal quantum computer with quantum speedup as presented in section~\ref{subsec:QBLAS-based_TF}, the fully coherent evolution and considerably high circuit depth are required, which is actually prohibited under current quantum hardware conditions. Thus, the implementation of the VQTF framework on the NISQ devices is presented in this section as a feasible alternative.  

\subsubsection{Variational quantum label predictor}
\label{subsubsec:VQLP}
Before the construction of the variational quantum information fusion channel, a multi-class classifier $f_{c}$ based on the NISQ devices~\citep{VQC}, is designed as Algorithm~\ref{alg2} for predicting the pseudo labels of $X_{t}$. In the VQTF framework, $f_{c}$ is invoked to be trained on $X_{s}$ and to classify $X_{t}$ into $C$ classes. 

In the training section, a binary classifier $f_{1}$ is firstly trained on the source domain data $X_{s}$ to separate the data of class $c_{1}$, any one class of the data set, from the remaining $C-1$ classes. Then, $f_{2}$ is trained on the remaining data to classify the second class from the temporary remaining data with $C-2$ different classes. Repeat this process until the $f_{C-1}$ classifier is trained on the remaining data to divide them into two classes with corresponding labels. After the above training process, the $C-1$ classifiers are applied to the target domain data in order of $f_{1}, f_{2}, \cdots, f_{C-1}$ to predict all the pseudo labels of the target data $X_{t}$.

For the $c$th class, the corresponding classifier 
\begin{equation}
	f^{(c)}(x; \theta^{(c)}, b^{(c)}) = \begin{cases}
		1&for \ \left \vert \textbf{U}_{\theta^{(c)}} \phi(x^{(c)}) \right \vert^{2} + b^{(c)} > \frac{n_{s}^{(c)}}{n_{s}}, \\ 
		0&for \ others,\end{cases}
	\label{eq:f_1}
\end{equation}
where 
\begin{equation}
	\textbf{U}_{\theta^{(c)}} = \textbf{U}_{L} \cdots \textbf{U}_{l} \cdots \textbf{U}_{1}
	\label{eq:U_theta_c}
\end{equation}	
represents the hierarchical quantum circuits with a series of parameters $\{\theta^{(c)}\}$ with the basic unitary layer $\textbf{U}_{l}$ for $l = \{ 1, \cdots, L \}$; $b_{c}$ is the corresponding bias; $n_{s}^{(c)}$ is the number of data points in class $c$. In order to avoid the class imbalance, the classifier threshold is set to be $n_{s}^{(c)} / n_{s}$.

Hence, the loss function of $f_{c}$ is 
\begin{eqnarray}
	\mathcal{L}_{c}(\theta^{(c)}, b^{(c)}) &= \frac{1}{2} \sum_{i=1}^{n_{s}^{(c)}} \left \vert \left \vert \textbf{U}_{\theta^{(c)}} \phi(x_{si}^{(c)}) \right \vert^{2} + b^{(c)} - y_{si}^{(c)} \right \vert^{2} \nonumber \\ 
	&= \frac{1}{2} \sum_{i=1}^{n_{s}^{(c)}} \left \vert P(1) + b^{(c)} - y_{si}^{(c)} \right \vert^{2},
	\label{eq:f_c loss function}
\end{eqnarray}
where $\{ x_{si}^{(c)}, y_{si}^{(c)} \}_{i=1}^{n_{s}^{(c)}} \in \mathcal{D}_{s}^{(c)}$; $P(1) = \left \vert \textbf{U}_{\theta^{(c)}} \phi(x_{si}^{(c)}) \right \vert^{2}$ is the successful probability of measurement operations. 

\begin{algorithm}[htbp]
	\caption{Variational quantum multi-class classifier}
	\KwIn{Source domain data $X_{s} = \{ x_{i}^{(s)} \}_{i=1}^{n_{s}}$ with labels $Y_{s} = \{ y_{i} \}_{i=1}^{(s)}$, target domain data $X_{t} = \{ x_{j}^{(t)} \}_{j=1}^{n_{t}}$}
	\KwOut{Target labels $\{ y_{j}^{(t)} \}_{j=1}^{n_{t}}$}
	\emph{step 1}: \\
	\For{$c = {1, \cdots, C}$}{
		Train the classifier $f^{(c)}(x; \theta^{(c)}, b_{c})$ on the data $X_{c} \vee X_{c + 1} \vee \cdots\vee X_{C}$. \\
	}
	Return the optimal classifiers $\{ f^{(c)}_{opt} \}_{c=1}^{C}$. \\
	\emph{step 2}: \\
	\For{$c = {1, \cdots, C}$}{
		Apply the trained classifiers $\{ f^{(c)}_{opt} \}_{c=1}^{C}$ on the target domain data $X_{t}$. \\
	}
	Return the target labels $\{ y_{j}^{(t)} \}_{j=1}^{n_{t}}$.
	\label{alg2}
\end{algorithm}

In addition, the parameters can be optimized by the gradient descent as follows
\begin{equation}
	\begin{cases}
		\theta_{t}^{(c)} = \theta_{t-1}^{(c)} - \alpha \frac{\partial \mathcal{L}_{c}(\theta^{(c)}, b)}{\partial \theta^{(c)}}, \\
		b_{t}^{(c)} = b_{t-1}^{(c)} - \alpha \frac{\partial \mathcal{L}_{c}(\theta^{(c)}, b)}{\partial b},
	\end{cases}
	\label{eq:parameters update}
\end{equation}
where $\alpha$ is the learning rate; $t$ is the number of the learning step. Ultimately, the optimal parameters $\theta_{opt}^{(c)}$, $b_{opt}^{(c)}$ can be obtained through the iterative optimization.

\subsubsection{Variational quantum transfer fusion}
\label{subsubsec:VQTF}
\begin{figure*}[t]
	\centering
	\includegraphics[width=0.9\textwidth]{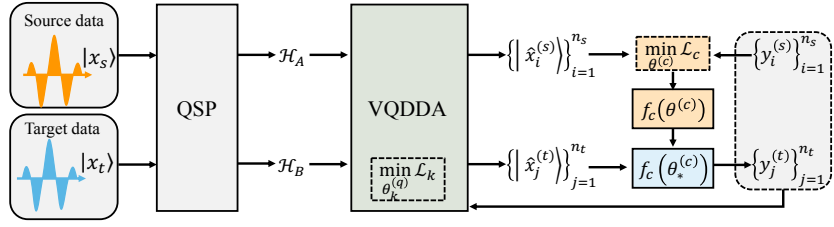}
	\caption{The illustrative diagram of the VQTF}
	\label{Fig:5}
\end{figure*}

The specific implementation of the VQTF is presented as follows. Inspired from~\citep{QGEP_1,QGEP_2,QGEP_3}, the whole process of the VQTF is depicted in Fig.~\ref{Fig:5}.

In step 1, prepare the initial state $\rho = | 0 \rangle \langle 0 |^{\otimes \log D}$. Let the Hamiltonians $\mathcal{H}_{A} = KL_{Q}K + \mu I$, $\mathcal{H}_{B} = KMK$.

In step 2, for $k = 1, 2, \cdots, d$, minimize the loss function 
\begin{equation}
	\mathcal{L}_{k}(\theta^{(q)}_{k}) = \frac{\mathrm{tr}[ \mathcal{H}_{A} \textbf{U}_{q}(\theta^{(q)}_{k}) \rho \textbf{U}^{\dagger}_{q}(\theta^{(q)}_{k})]}{\mathrm{tr}[\mathcal{H}_{B} \textbf{U}_{q}(\theta^{(q)}_{k}) \rho \textbf{U}^{\dagger}_{q}(\theta^{(q)}_{k})]} + \sum_{i=1}^{k-1}\alpha_{i} \mathrm{tr}[\mathcal{H}_{B} \textbf{U}_{q}(\theta^{(q)}_{k}) \rho \textbf{U}^{\dagger}_{q}(\theta^{(q)}_{k})]^{2} 
	\label{eq:L_Q}
\end{equation}
where $\textbf{U}_{q}(\theta^{(q)}_{k})$ can be constructed by the layered quantum circuits with a series of parameters $\{ \theta^{(q)}_{k} \}_{k=1}^{d}$; $\alpha_{i}$ is a coefficient to balance the two terms of the loss. The first term is designed for the generalized eigen-problem. The second term is actually a regularized restriction for avoiding overfitting. 

Based on the method of gradient descent, the process of optimization starts with $\mathcal{L}_{1}$ whose minimum results in the ground state $| \psi(\theta^{(q)}_{1}) \rangle$ of the generalized eigenproblem of Eq.~\eqref{eq:L_Q}. Subsequently, $| \psi(\lambda_{1}) \rangle$ is substituted into the cost function $\mathcal{L}_{2}(\theta^{(q)}_{2})$ to obtain the second excite state $| \psi(\theta^{(q)}_{2}) \rangle$. Iterate this process and the $k$th eigenstate $| \psi(\theta^{(q)}_{k}) \rangle$ can be obtained by substituting the $| \psi(\theta^{(q)}_{1}) \rangle, \cdots, | \psi(\theta^{(q)}_{k-1}) \rangle$ to $\mathcal{L}_{k}(\theta^{(q)}_{k})$ and achieving its minimum. Hence, the transformation matrix $W$ can be acquired by the $k$ eigenvectors corresponding to the $k$ smallest eigenvalues of Eq.~\eqref{eq:L_Q}. 

Then, the quantum classifiers $\{f_{c}\}_{c=1}^{C}$ as designed in section~\ref{subsec:VQLP} are applied to the transformed data $\hat{X} = W^{T} X$ to predict the pseudo target labels $\{ \hat{y}_{j}^{(t)} = f(W^{T} x_{j}^{(t)}) \}_{j=1}^{n_{t}}$. The quantum multi-class predictor is firstly trained on $\hat{X}_{s}$ and then applied to $\hat{X}_{t}$ to predict the labels. Repeat the above steps until the convergence. The target labels $Y_{t} = \{ y_{i}^{(t)} \}_{j=1}^{n_{t}}$ can be ultimately obtained. The algorithm pseudo-code of the VQTF is presented in Algorithm~\ref{alg3}.

\begin{algorithm}
	\caption{VQTF framework}
	\KwIn{$\mathcal{H}_{A} = KL_{Q}K + \mu I$, $\mathcal{H}_{B} = KMK$}
	\KwOut{Target labels $\{ y_{j} \}_{j=1}^{n_{t}}$}
	\emph{step 1}: Prepare the ansatz state $\{ | \psi(\lambda_{k}) \rangle \}_{k=1}^{d}$ by a parameterized quantum circuit with a set of parameters $\{ \theta_{p}^{(a)} \}$. \\
	\emph{step 2}: \\
	\Repeat{Convergence}{
	\For{$k = 1, \cdots, d$}{
		(1) Compute the cost function $\mathcal{L}_{k}$ in Eq.~\eqref{eq:L_Q}; \\
		(2) Optimize the cost function to the minimum with the gradient descent algorithm to obtain the transformation matrix $W$; \\
	}
	
	(3) Compute the new data $\hat{X} = W^{T}X = [\hat{X}_{s}, \hat{X}_{t}]$; \\
	(4) Invoke the quantum classifier $\{f_{c}\}_{c=1}^{C}$ in section~\ref{subsubsec:VQLP} to update the pseudo target labels $\{ \tilde{y}_{j}^{(t)} = f(W^{T} x_{j}^{(t)}) \}_{j=1}^{n_{t}}$. \\
	}
	Return the target labels $Y_{t} = \{ y_{j}^{(t)} \}_{j=1}^{n_{t}}$.
	\label{alg3}
\end{algorithm}

\section{Numerical experiments}
\label{sec:numerical experiments}
In this section, the numerical experiments for demonstrating the feasibility and effectiveness of the VQTF are implemented by comparing the performance of the corresponding algorithms on different types of datasets. All the algorithms in the experiments are simulated on the classical computer in Python language. The gradient descent based optimizer is Adam~\citep{ADAM}. In the following, the settings of the datasets, the benchmark models, and the results analysis of the numerical experiments are presented respectively.

\subsection{Datasets}
\label{subsec:datasets}
\begin{figure*}
	\centering
	\begin{subfigure}{0.45\textwidth}
	\centering
	\includegraphics[width=\textwidth]{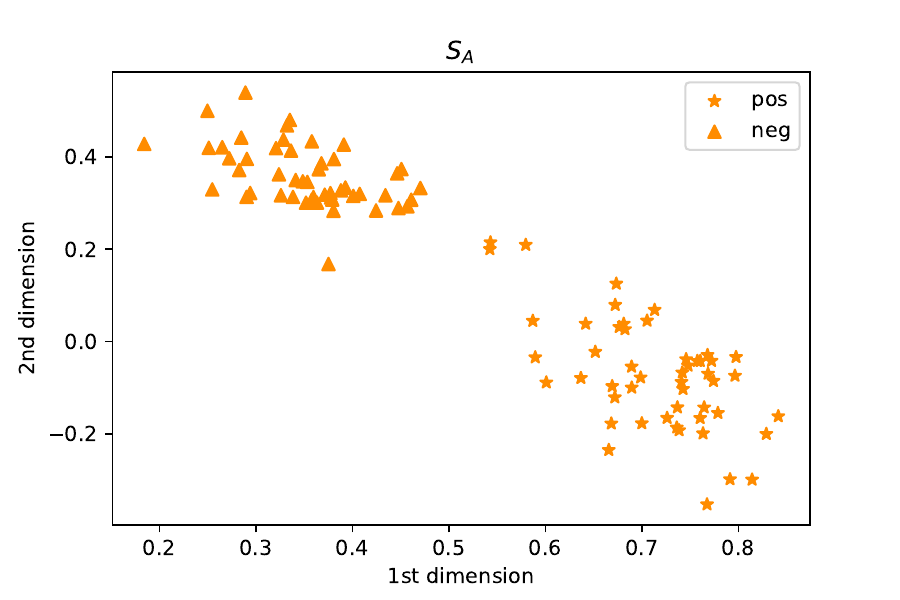}
	\caption{}
	\label{Fig.6a}
	\end{subfigure}	
	\begin{subfigure}{0.45\textwidth}
	\centering
	\includegraphics[width=\textwidth]{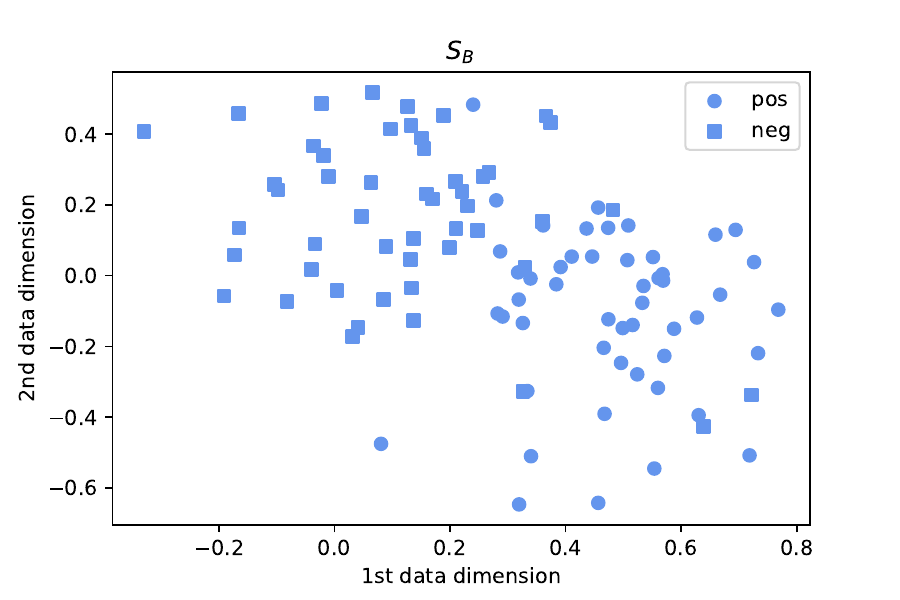} 
	\caption{}
	\label{Fig.6b}
	\end{subfigure}
	\\
	\begin{subfigure}{0.45\textwidth}
	\centering
	\includegraphics[width=\textwidth]{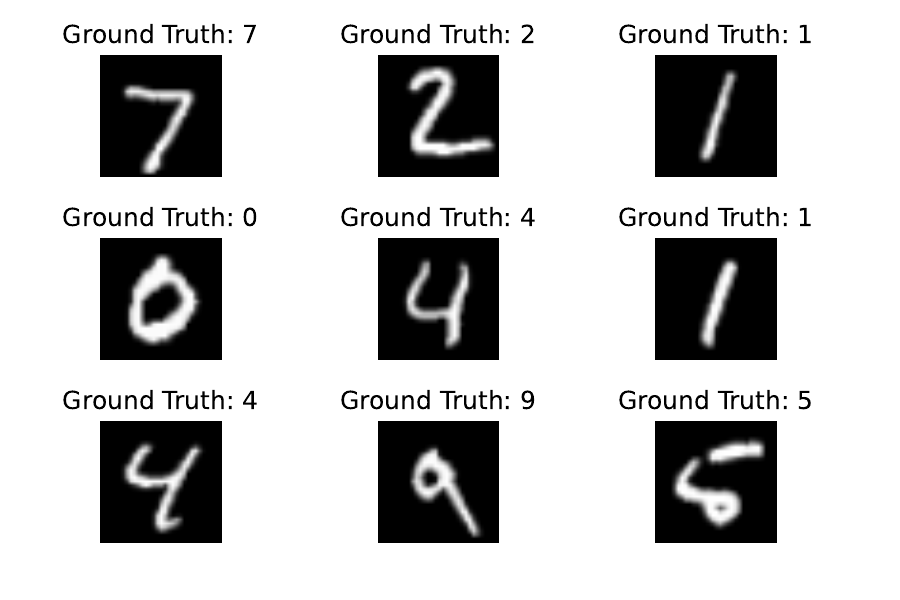}
	\caption{}
	\label{Fig.6c}
	\end{subfigure}	
	\begin{subfigure}{0.45\textwidth}
	\centering
	\includegraphics[width=\textwidth]{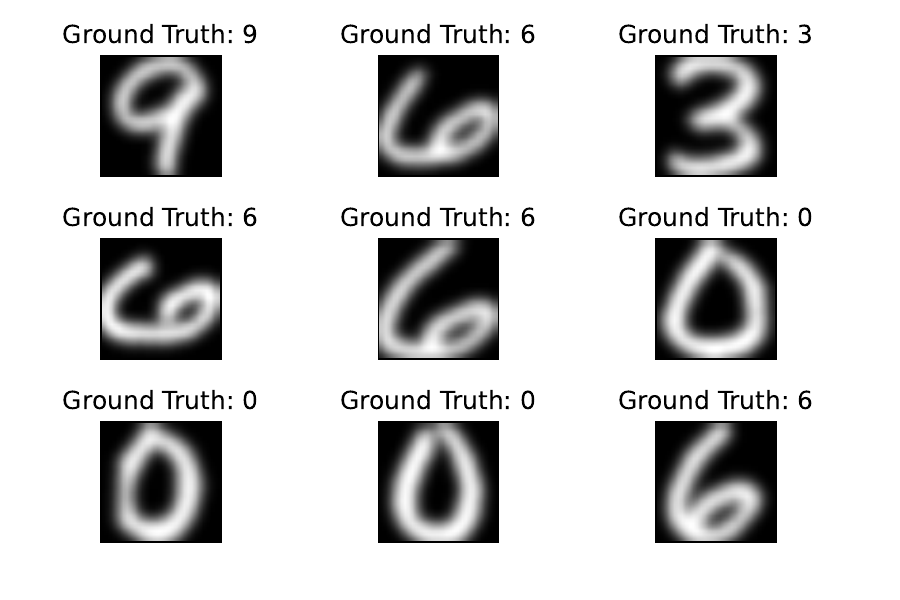} 
	\caption{}
	\label{Fig.6d}
	\end{subfigure}
	\caption{The visualization of the datasets. (a)the synthetic $S_{A}$ data; (b)the synthetic $S_{B}$ data; (c)the MNIST data; (d)the USPS data.}
	\label{Fig:6}
\end{figure*}

The synthetic dataset is created to simulate the 2-qubit quantum data scenario. This dataset mainly contains two domains, namely $S_{A}$ and $S_{B}$, generated from different distributions respectively. Specifically, the domain $S_{A} = \{ | x_{i}^{(s)} \rangle \}_{i=1}^{100} \in \mathbb{R}^{4}$ are randomly normalized samples from the distribution $\mathcal{N}(\mu_{1}^{A}=\mu_{2}^{A}=0, \sigma_{1}^{A}=\sigma_{2}^{A}=1)$ as depicted in Fig.~\ref{Fig.6a}; the domain $S_{B} = \{ | x_{j}^{(t)} \rangle \}_{i=1}^{100} \in \mathbb{R}^{4}$ are randomly normalized samples from the distribution $\mathcal{N}(\mu_{1}^{B}=\mu_{2}^{B}=1, \sigma_{1}^{B}=\sigma_{2}^{B}=2)$ as depicted in Fig.~\ref{Fig.6b}. Both domains contain 100 samples evenly in two different classes. 

In addition to the synthetic dataset, the handwritten digits dataset, the MNIST~\citep{MNIST} and the USPS~\citep{USPS}, are selected alternatively as the source and target domain data. The data samples from the two domains both contains 10 classes, namely the handwritten digits from 0 to 9 as sampled in Fig.~\ref{Fig.6c} and Fig.~\ref{Fig.6d}. But the samples of the MNIST and the USPS have differences in angles and background. In our experiment, 2000 images in size of $28*28$ are sampled from the MNIST; 1800 images in size of $16*16$ are sampled from the USPS. 

\subsection{Benchmark models}
\label{subsec:models}
To demonstrate that the VQTF can achieve the SOTA performance among the current variational quantum DA algorithms, namely the VQTCA, the VQCORAL, the VQSA, and the VQDAC, are selected as the benchmark models. Furthermore, the classical DDA algorithms such as the TCA, the JDA, and the BDA, are reproduced as the comparison to exhibit that our method can achieve at least competitive performance. In addition, the no adaptation model (NA), which is trained on the source domain and directly transferred to the target domain without any DA techniques, is also added as the baseline. All the benchmark models are performed on the two datasets above resulting in the results presented in Table~\ref{tab:accuracy}.

\subsection{Results analysis}
\label{subsec:results}
In Table~\ref{tab:accuracy}, the TL task is represented as $\mathcal{D}_{s} \rightarrow \mathcal{D}_{t}$. According to the results, the VQTF proposed in this paper can achieve competitive performance in the numerical experiments. From the results on the synthetic dataset, it seems that quantum feature alignment methods such as the VQCORAL, the VQSA, and the VQDAC may not be efficient for the task of distribution domain adaptation. Although the superiority of the VQTF is not clear from the prediction accuracy on the synthetic data compared to the classical methods, the data are assumed to be quantum states. It means that the classical methods need the quantum state tomography operations, which are extremely time-consuming to extract the feature information from quantum data. Thus, the quantum advantage of the VQTF can be guaranteed under this circumstance with comparable performance. Overall, the results from the numerical experiments shows that the VQTF can deal with DA tasks feasibly and effectively.

\begin{table}[t]
	\caption{\label{tab:accuracy}%
	Prediction accuracy of different methods on the synthetic datasets and the handwritten digits datasets.
	}
	\resizebox{\textwidth}{!}
	{
	\begin{tabular}{ccccc}
		\toprule
		\textrm{} & \textrm{$S_{A} \rightarrow S_{B}$} & \textrm{$S_{B}\rightarrow S_{A}$} & \textrm{MNIST $\rightarrow$ USPS} & \textrm{USPS $\rightarrow$ MNIST}\\
		\hline
		NA & 44.42\% & 49.87\% & 64.40\% & 35.91\%\cr
		TCA~\citep{TCA} & 88.19\% & 85.78\% & 54.30\% & 52.10\%\cr
		JDA~\citep{JDA} & 97.64\% & 97.36\% & 62.42\% & 56.99\%\cr
		BDA~\citep{BDA} & 98.99\% & 96.28\% & 65.78\% & 57.68\%\cr
		VQTCA~\citep{QTCA} & 70.33\% & 66.78\% & 53.35\% & 50.01\%\cr
		VQCORAL~\citep{QCORAL} & 14.11\% & 4.05\% & 65.64\% & 44.55\%\cr
		VQSA~\citep{QSA} & 7.66\% & 10.21\% & 64.90\% & 36.04\%\cr
		VQCDA~\citep{QCDA} & 13.68\% & 5.56\% & 65.60\% & 44.60\%\cr
		VQTF & 99.05\% & 97.44\% & 66.66\% & 57.87\%\cr
		\bottomrule
	\end{tabular}
	}
\end{table}

\section{Conclusion}
\label{sec:conclution}
In this paper, the procedure of TL is achieved on quantum devices by aligning the data distributions of the source and target domain. In terms of the type of the quantum device, the transfer infusion procedure is implemented on the universal quantum computer and the NISQ devices respectively. The former implementation can achieve quantum speedup by the QPE and conditional rotation operations based on the sparsity of the given data in the kernel space. The latter implementation can achieve competitive performance compared to the SOTA QDA methods, and quantum advantage in the quantum data scenario.

However, many open problems still exist in both implementations. For instance, the implementation of the QBLAS-based TF on the hardware needs to decompose the advanced module as presented in this paper to basic one-qubit and two-qubit operations. The underlying circuit depth may be prohibited with the increase of the data scale and the requirement of the prediction accuracy. Thus, how to authentically realize the theoretical quantum speedup needs much more exploration. In addition, the quantum advantage of the VQTF on classical problems is still theoretically ambiguous. The influence of the data scale and the noise on the performance of the VQTF are also unclear. In terms of quantum learning theory, the generalization ability and the quantum effectiveness of the QTL method also should be bounded for guidance. The QTL method is actually a completely new sub-realm of the QML. It can be specialized for the problem of data label scarcity and cross-domain learning. Our work is hoped to encourage much more explorations to the field of QTL.  

\section*{Acknowledgements}
The author would like to thank the referees for helpful comments on this paper. This work is supported by National Key Research and Development Program of China Grant No. 2018YFA0306703, in part by the National Natural Science Foundation of China under Grant 62271296, in part by Natural Science Basic Research Program of Shaanxi (No. 2021JC-47), in part by Key Research and Development Program of Shaanxi (Program No. 2022GY-436, NO. 2021ZDLGY08-07), and in part by the Natural Science Basic Research Program of Shaanxi (No. 2022JQ-018), in part by the Natural Science Foundation of Sichuan Province (Grant No. 2023NSFSC1068).


\bibliographystyle{elsarticle-num} 
\bibliography{DATF.bib}





\end{document}